\pdfoutput=1

\documentclass[utf8]{FrontiersinHarvard} 

\usepackage{url,hyperref,lineno,microtype,subcaption}
\usepackage[onehalfspacing]{setspace}
\usepackage{pifont}
\newcommand{\cmark}{\ding{51}}%
\newcommand{\xmark}{\ding{55}}%

\def\keyFont{\fontsize{8}{11}\helveticabold }
\def\firstAuthorLast{Fan {et~al.}} 
\def\Authors{Zhipeng Fan\,$^{1}$, Yao Wang\,$^{1,2,*}$}
\def\Address{$^{1}$Department of Electrical and Computer Engineering, Tandon School of Engineering, New York University, Brooklyn, NY, USA\\
$^{2}$Department of Biomedical Engineering, Tandon School of Engineering, New York University, Brooklyn, NY, USA}
\def\corrAuthor{Yao Wang}

\def\corrEmail{yaowang@nyu.edu}

\begin{document}
\onecolumn
\firstpage{1}

\title[Lite I2L Mesh Net]{Lightweight Estimation of Hand Mesh and Biomechanically Feasible Kinematic Parameters} 


\author[\firstAuthorLast ]{\Authors} 
\address{} 
\correspondance{} 

\extraAuth{}

\maketitle

\begin{abstract}

\section{}

3D hand pose estimation is a long-standing challenge in both robotics and computer vision communities due to its implicit depth ambiguity and often strong self-occlusion. 
Recently, in addition to the hand skeleton, jointly estimating hand pose and shape has gained more attraction. 
State-of-the-art methods adopt a model-free approach, estimating the vertices of the hand mesh directly and providing superior accuracy compared to traditional model-based methods directly regressing the parameters of the parametric hand mesh. 
However, with the large number of mesh vertices to estimate, these methods are often slow in inference. 
We propose an efficient variation of the previously proposed image-to-lixel approach to efficiently estimate hand meshes from the images. Leveraging recent developments in efficient neural architectures, we significantly reduce the computation complexity without sacrificing the estimation accuracy.
Furthermore, we introduce an inverted kinematic(IK) network to translate the estimated hand mesh to a biomechanically feasible set of joint rotation parameters, which is necessary for applications that leverage pose estimation for controlling robotic hands. 
Finally, an optional post-processing module is proposed to refine the rotation and shape parameters to compensate for the error introduced by the IK net. 
Our Lite I2L Mesh Net achieves state-of-the-art joint and mesh estimation accuracy with less than $13\%$ of the total computational complexity of the original I2L hand mesh estimator. Adding the IK net and post-optimization modules can improve the accuracy slightly at a small computation cost, but more importantly, provide the kinematic parameters required for robotic applications.

\tiny
 \keyFont{ \section{Keywords:} computer vision, 3D hand pose estimation, efficient models, biomechanically feasible kinematics estimation, 3D hand fitting} 
\end{abstract}

\section{Introduction}

3D hand pose estimation is a long-standing challenge in computer vision communities and has seen growing interest recently due to its wide-range applications in robotics, AR/VR, and human-computer interaction (HCI) (\cite{sridhar2015investigating}). The ability to accurately estimate hand poses could enable a wide range of applications, such as virtual grasp(\cite{chessa2019grasping}), control of a robotic hand, or even telesurgery. However, because of the implicit depth ambiguity and the often severe (self-)occlusions, estimating the 3D positions of the hand joints with high precision remains a highly active research area.

Over the years, many works have been proposed to address the 3D hand pose estimation problem (\cite{zimmermann2017learning, iqbal2018hand, ge2018hand, cai2018weakly, zimmermann2019freihand, zhou2020monocular}). Traditional methods either regress the 3D joint coordinates directly or discretize the 3D space encircling the hand and predict the heatmap corresponding to the probabilities of the joints at each position in the space. These joint-based pose representations are relatively efficient to estimate because there are usually fewer than 20 joints defined on the hand. However, the joint position information may not be sufficient in many applications. For example, knowing only the skeleton of the hand is not enough to determine the surface of the hand and whether it is in contact with the object, which is vital for many applications like realistic AR/VR effects (\cite{wan2004realistic}) or touch processing (\cite{lambeta2021pytouch}). 

Therefore, more recently, many works have adopted parametric hand models (e.g. MANO \cite{MANO:SIGGRAPHASIA:2017}) to jointly estimate the hand shape and hand poses. Most of the work directly predicts the rotation parameters of the hand models and then deforms the rigged hand mesh using linear blend skinning. The resulting mesh is then compared with the ground-truth mesh obtained through fitting to compute the loss. More recently, some works (\cite{ge20193d, moon2020i2l, lin2021end, lin2021mesh}) propose to estimate the vertices of the hand mesh directly from images and show that this greatly improves the accuracy of pose estimation compared to parameter regression-based methods. This is because it is easier to localize the points of interest from the 2D image than learning the implicit and highly nonlinear axis-angle rotation parameters used by the existing parametric hand model. However, these methods tend to be computationally heavy, since significantly more points of interest are estimated, which calls for more sophisticated backbone models such as stacked ResNet (\cite{xiao2018simple}) and HRNet (\cite{sun2019deep}) as well as decoder modules such as GCN (\cite{ge20193d}) and Transformer (\cite{lin2021end, lin2021mesh}). These heavy computations prohibit the usage of such models on mobile devices for real-time applications.

To improve the real-time performance of such models, we examine the computational profile of one of the state-of-the-art hand pose estimation models, the I2L Mesh Net (\cite{moon2020i2l}) and propose a lighter version of it, motivated by recent progress in efficient deep learning models. Furthermore, we propose an efficient decoder module that works in conjunction with the encoder. To further improve the performance of the pose estimator, we introduce a pipeline to generate realistic rendering data as additional training data. The pipeline requires very little human intervention and could greatly improve the accuracy of the light weight image to lixel mesh net, allowing it to obtain state-of-the-art performance with less than $13\%$ computational costs.

The estimated hand meshes could be transferred back to the parametric space of the hand model for many downstream tasks, such as robotic hand control, motion re-targetting, etc. This process of transferring from the resulting end joint positions back to the rotation parameters is generally termed as inverse kinematics (IK) (\cite{kucuk2006robot}). Compared to the Forward Kinematic (FK) process, where the end positions of each joint are estimated based on the given rotation parameters, the IK process is more difficult to solve due to the nonlinearity. We introduce a biomechanically inspired inverse kinematic (IK) process leveraging the statistical human hand model MANO (\cite{MANO:SIGGRAPHASIA:2017}). Unlike the traditional design of the IK net, which has 45 rotation parameters, we restrict the space of the rotation parameters to be biomechanically feasible, with 23 parameters following \cite{lim2020mobilehand} to assist the downstream robotic applications, for example, to control the movement of a robotic. To properly train this bioinspired IK net, we generate ground truth rotation parameters by an iterative fitting process, which minimizes the location difference of joints and mesh vertices corresponding to the original hand mesh controlled by the full set of 45 rotation parameters and those corresponding to the mesh controlled by our 23 biomechanically feasible parameters. Finally, following recent work (\cite{kolotouros2019learning, lv2021handtailor}), we additionally introduce a light post-optimization module, which further encourages the resulting hand mesh from the IK module to match the one estimated by our lite I2L mesh net. 

The remainder of this paper is organized as follows. We will review the literature in the related work section and then introduce the various modules of our pipeline in more detail in the Methods section. In the Experiment section, we will provide more details of our implementation and illustrate our experimental results. Finally, we will summarize the main contributions of this work in the conclusion section.

\section{Related Work}

\subsection{Hand Pose Estimation}
3D hand pose estimation is an evergreen topic in the computer vision domain due to its wide range of applications yet inherent challenges. The pioneering works (\cite{rehg1994digiteyes, lu2003using, stenger2006model, oikonomidis2011full}) adopt optimization-based methods for hand fitting, which are often slow and pruned to error. With the rapid developments of deep learning, most of the recent work adopts deep neural nets to perform pose estimation. 

Traditional models that based on deep learning for hand pose estimation, including \cite{zimmermann2017learning,oberweger2017deepprior++, mueller2018ganerated,spurr2018cross,iqbal2018hand,cai2018weakly,ge2018hand}, focus on estimating the 3D layout of the skeleton, which is equivalent to estimate the 3D coordinates of the hand joints. 
More specifically, \cite{zimmermann2017learning} propose a deep neural network to learn the 3D pose priors from images and combine the learned prior with the estimated keypoints to generate the final 3D poses. \cite{mueller2018ganerated} introduce a synthetic dataset and use GAN (\cite{goodfellow2014generative}) to reduce the domain gap. \cite{spurr2018cross} propose a Variational Autoencoder (VAE) based methods to align the latent pose space of models trained on synthetic data and nature RGB images. Instead of the absolute depth, the root relative depth was estimated by \cite{iqbal2018hand} to eliminate the ambiguity of the depth. A weakly supervised method that takes advantage of the depth map as additional supervision was proposed by \cite{cai2018weakly}. These methods only estimate the skeleton structure of the hand without the surface information, which may not be sufficient for applications with higher precision demands, such as controlling of robotic hand, virtual jewelry try-on, etc.

Most of the recent work jointly estimates the hand pose and the hand shape from the RGB inputs. Statistical parametric hand models like MANO \cite{MANO:SIGGRAPHASIA:2017} are often used to describe hand shapes. Specifically, model-based methods (\cite{hasson2019learning,zhang2019end, boukhayma20193d, zimmermann2019freihand, yang2020bihand, zhou2020monocular, liu2021semi, lv2021handtailor, chen2021model, lim2020mobilehand}) estimate the pose of the hand by predicting the rotation parameters of the MANO model. The mapping between the RGB images and these rotation parameters is highly abstract and poses great obstacles to network learning. More recently, model-free methods have become more popular. \cite{ge20193d} and \cite{chen2022mobrecon} use the graph convolution network (GCN) to regress the coordinates of the mesh vertices directly from the images. However, the direct regression-based approach often has a lower accuracy and lacks a way to model the uncertainties \cite{li2021human}. \cite{moon2020i2l} predicts the coordinates of 3D vertices coordinates as 3 1D lixel heatmaps, greatly reducing computational and memory cost. However, despite the effort to reparameterize the heatmap representations, the computational cost remains high due to the usage of computationally heavy encoders and decoders. \cite{lin2021end} and \cite{lin2021mesh} further adopt HRNet (\cite{sun2019deep}) and Transformer (\cite{vaswani2017attention}) to regress the 3D coordinates, which are even more computationally heavy and are not suitable for mobile applications. 

In this work, we propose a lite version of the I2L Mesh Net (\cite{moon2020i2l}). By replacing the encoder model and the decoder module with more efficient variants, we greatly reduce the computational burden of the I2L Mesh Net without sacrificing the advantages of the I2L Mesh Net, i.e., preserving the spatial relationships between the input pixels and the mesh vertices, allowing easier modeling of the prediction uncertainties (\cite{moon2020i2l}).

\subsection{Efficient Models}

With the ubiquitous presence of mobile devices, adapting deep neural networks for deployment on mobile devices has gained increasing interest in recent years. 
SqueezeNet (\cite{iandola2016squeezenet}) develops channel squeeze and expansion operations to improve the efficiency of the model. \cite{gholami2018squeezenext} further extend the SqueezeNet~V1 to obtain better accuracy and efficiency trade-offs by using more aggressive squeeze operations and separable convolutions. 
Depthwise separable convolutions were first proposed in MobileNet~V1 (\cite{howard2017mobilenets}) to reduce computational costs. In MobileNet~V2 (\cite{sandler2018mobilenetv2}), inverted residual was proposed to first expand the number of channels for feature extraction, followed by channel squeeze to reduce channel dimensionality. This design improves the capacity of the model with limited resources. Finally, in the latest version (\cite{howard2019searching}), neural architecture search (NAS) is adopted to find more efficient network architectures. 
Another line of work, the ShuffleNet series (\cite{zhang2018shufflenet, ma2018shufflenet}) use group convolution and channel shuffle to improve efficiency. Furthermore, the memory access cost is also considered to further reduce the inference time. 
The EfficientNet (\cite{tan2019efficientnet}) takes a different path and investigates the proper way to scale up an efficient backbone. Similarly, NAS was used to find the most efficient backbone architecture, while for the basic building block, the inverted bottleneck MBConv from MobileNet~V2 was adopted. The EfficientNet~V2 further adds the minimization of the training time into the NAS objectives and proposes an efficient model that balances the resource consumption, the accuracy, and the training speed.

Our lite I2L Mesh Net is built on these prior efforts in developing efficient models. We adopt EfficientNet (\cite{tan2019efficientnet}) as the backbone for feature extraction and introduce an efficient decoding module to pair with the backbone, to achieve a good balance between the pose estimation accuracy and the model complexity.

\section{Method}
\begin{figure}[h!]
\begin{center}
\includegraphics[width=\linewidth]{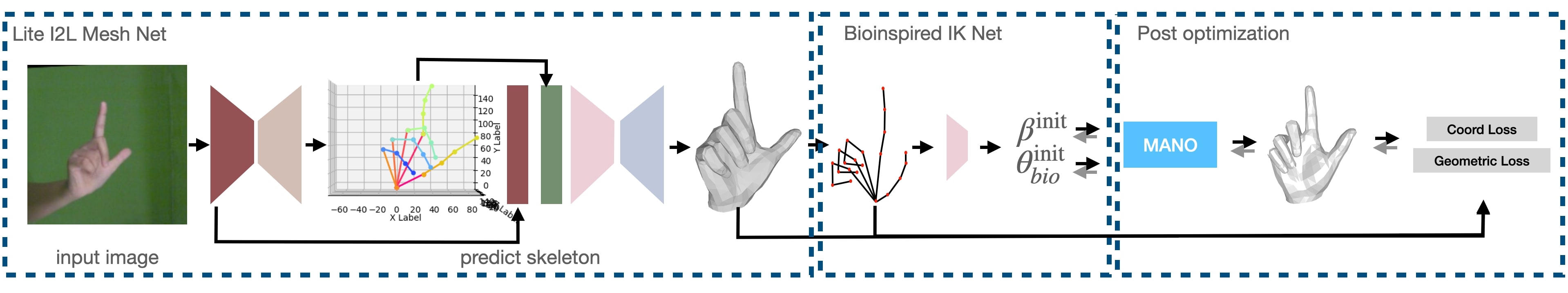}
\end{center}
\caption{The complete pipeline of our lite I2L Mesh Net, the bioinspired IK Net, and our post-optimization module. The lite I2L Mesh Net efficiently estimates the hand skeleton and the meshes. Then, from the estimated mesh, the corresponding skeleton is extracted and fed into the bioinspired IK Net to estimate the corresponding kinematic parameters.  }\label{fig:full_pipeline}
\end{figure}

The overall architecture of the original I2L mesh net is illustrated in Fig.~\ref{fig:pipeline}. Our pipeline to predict the initial hand mesh follows the same general structure as the I2L mesh net with a more efficient encoder and decoder design. We further introduce a bioinspired Inverse Kinematic (IK) net to estimate the biomechanically correct rotations at the joints, followed by an optional post-optimization module to fine-tune the predictions. The overall pipeline of the proposed system is shown in Fig. \ref{fig:full_pipeline}.

Given a cropped image of the hand, the image-to-lixel net estimates the canonical joint and vertice positions. The estimated skeleton is then sent to the IK module to predict the biomechanically feasible set of rotation parameters. To further remedy the potential error and infeasible poses in the estimated bone rotations, the final optional post-optimization module aims to find a better fit with the initial mesh prediction by iteratively tuning the rotation parameters. We present each module in more detail in the following sections.

\subsection{MANO: parametric hand mesh}

We first briefly introduce the MANO hand mesh, which was built from a human hand scan dataset containing more than 1000 high-resolution scans from 30 subjects. It captures a large variety of human-hand poses and also provides vivid geometry details. Specifically, MANO generates the set of vertices $\mathcal M$ of the hand mesh based on the shape parameters $\beta\in \mathbb{R}^{10}$ and the pose parameters $\theta \in \mathbb{R}^{45+3}$ (45 articulation parameters that describe the rotations at the internal joints and 3 global parameters describing the rotations of the root wrist joint. The resulting mesh could be computed as follows:
\begin{align}
    \mathcal{M}(\theta, \beta) &= \mathcal{W}(\mathcal{T}(\theta,\beta), \mathcal{J}(\beta), w)\\
    \mathcal{T}(\theta, \beta) &= \bar{T}+\mathcal{B_S}(\beta)+\mathcal{B_P}(\theta) 
\end{align}
where $\bar{T}$ denotes the average shape (i.e., the vertex coordinates of an average hand in a canonical rest pose), while $\mathcal{B_S}(\beta)$ and $\mathcal{B_P}(\theta)$ denote the offset from the mean shape for a specific hand, due to the shape variation described by $\beta$ and the pose variation captured by $\theta$. $\mathcal{J}(\beta)$ denotes the joint layout of the hand based on the specific hand shape $\beta$. Finally, $w$ is the linear blend weight between $\mathcal{M}$ and $\mathcal{J}$.

\subsection{Lite I2L Mesh Net}

In this section, we introduce the modifications we made to make the I2L Mesh Net more efficient. Specifically, we analyze the computational costs of each component of the original I2L Mesh Net and then replace them with either modified existing efficient backbones or newly proposed architectures. We will first provide a brief review of the I2L Mesh Net, followed by introducing our proposed light version. 

\begin{figure}[h!]
\begin{center}
\includegraphics[width=\linewidth]{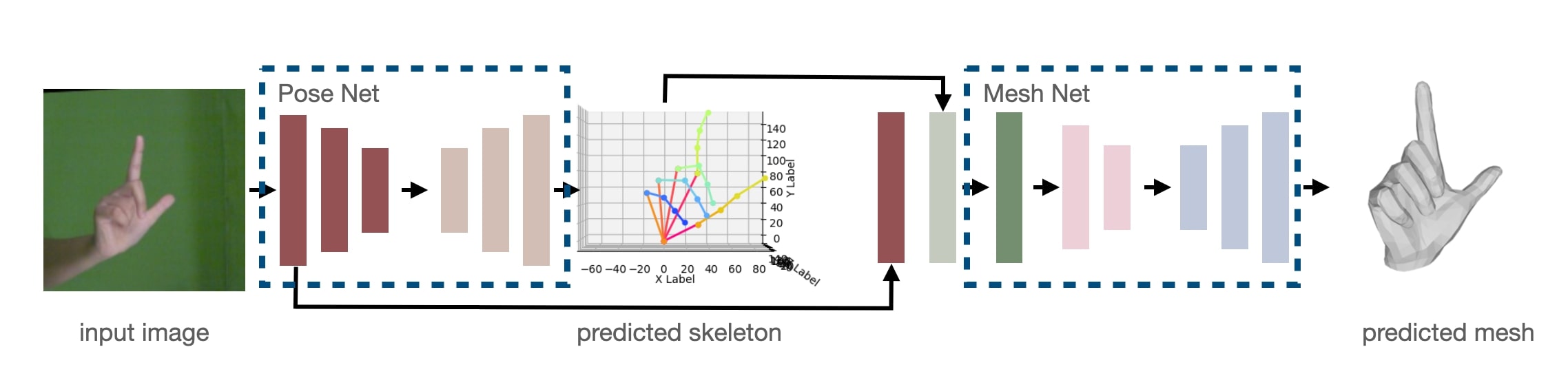}
\end{center}
\caption{The pipeline of the original I2L Mesh Net (\cite{moon2020i2l}). In our lite version, we adopt a similar pipeline to predict the hand mesh, but introduce more efficient modules for each component.}\label{fig:pipeline}
\end{figure}

\subsubsection{Review of I2L Mesh Net}

The I2L Mesh Net is one of the pioneer works to directly predict the 1D lixel (line+pixel) likelihood for each vertices using a heat map representation. Compared to the traditional rotation parameter regression approach, which directly learns the highly nonlinear mapping between rotations at the joints and the input images, the 1D heatmap representation better preserves the spatial relationships and also incorporates the potential uncertainty of the keypoint locations. As a result, better joint estimation accuracy is achieved. 

More specifically, the I2L Mesh Net adopts a stacked backbone, as shown in Fig. \ref{fig:pipeline}. The first stage (i.e. PoseNet) estimates the 3D coordinates for the hand joints, similar to the traditional encoder-decoder based pose estimation models (\cite{newell2016stacked, xiao2018simple}). The encoder uses the ResNet50 model architecture. The second stage (i.e. MeshNet) combines the early image features as well as the 3D pose in the 3D joint heatmap representation and uses another encoder-decoder structure to generate vertex heatmaps. Predicting the vanilla 3D heatmap for each vertex is computationally infeasible and memory heavy, therefore, the MeshNet decomposes the 3D heatmap to three 1D lixel-based heatmap describing the likely X-, Y-, and Z- coordinates of the vertex. To predict the two 1D heatmaps in the image plane (X and Y), the 2D features are marginalized over the Y and X directions, respectively. To predict the heatmap in the depth (Z) direction, the MeshNet uses global pooling to generate 1D feature maps for depth estimation.

We decompose the complete I2L Mesh Net into several individual functional components and analyze the computation cost for each. For the PoseNet, we denote its feature extraction part as the pose backbone and its heat map decoding part as the pose decoder. Similarly, we denote the corresponding part in MeshNet as the mesh backbone and the mesh decoder. Additionally, we denote the module to prepare features from joint-based heatmap representations as heatmap aggregation module. Assuming that the input resolution is $256\times256$, we then measure the computational cost for each module in terms of the multiply-accumulate operation (MAC) using the publicly available tool \href{https://github.com/sovrasov/flops-counter.pytorch}{flops-counter.pytorch} and present the results in the Tab. \ref{tab:i2l_flops}. The majority of the computation comes from the backbone and the decoder. The ResNet50-based backbone in both the PoseNet and the MeshNet costs about 5.3 GMACs (1 GMACs = $10^9$ MACs). The Decoder, although having a simpler structure, contains more computations. The majority of the computations in the decoder come from the feature map upsampling process. 

\begin{table}[t]
\centering
\small
\caption {Computational cost of the original I2L Mesh Net, when the input image resolution is $256\times256$ and the lixel heatmap resolution is $64$ for both the heatmap decoder and mesh decoder. The hand model contains 21 joints and 778 mesh vertices. } 
\begin{tabular}{lccccc}
\label{tab:i2l_flops}
\\\hline
Module  &  Pose Backbone & Pose Decoder &  Heatmap Aggregation & Mesh Backbone    & Mesh Decoder \\\hline
Cost (GMACs) &    5.38      & 7.55         &  3.32     & 5.22      & 7.59\\
\hline
\end{tabular}
\end{table}

\subsubsection{Simplifying the original I2L Mesh Net}
\begin{figure}[h!]
\begin{center}
\includegraphics[width=\linewidth]{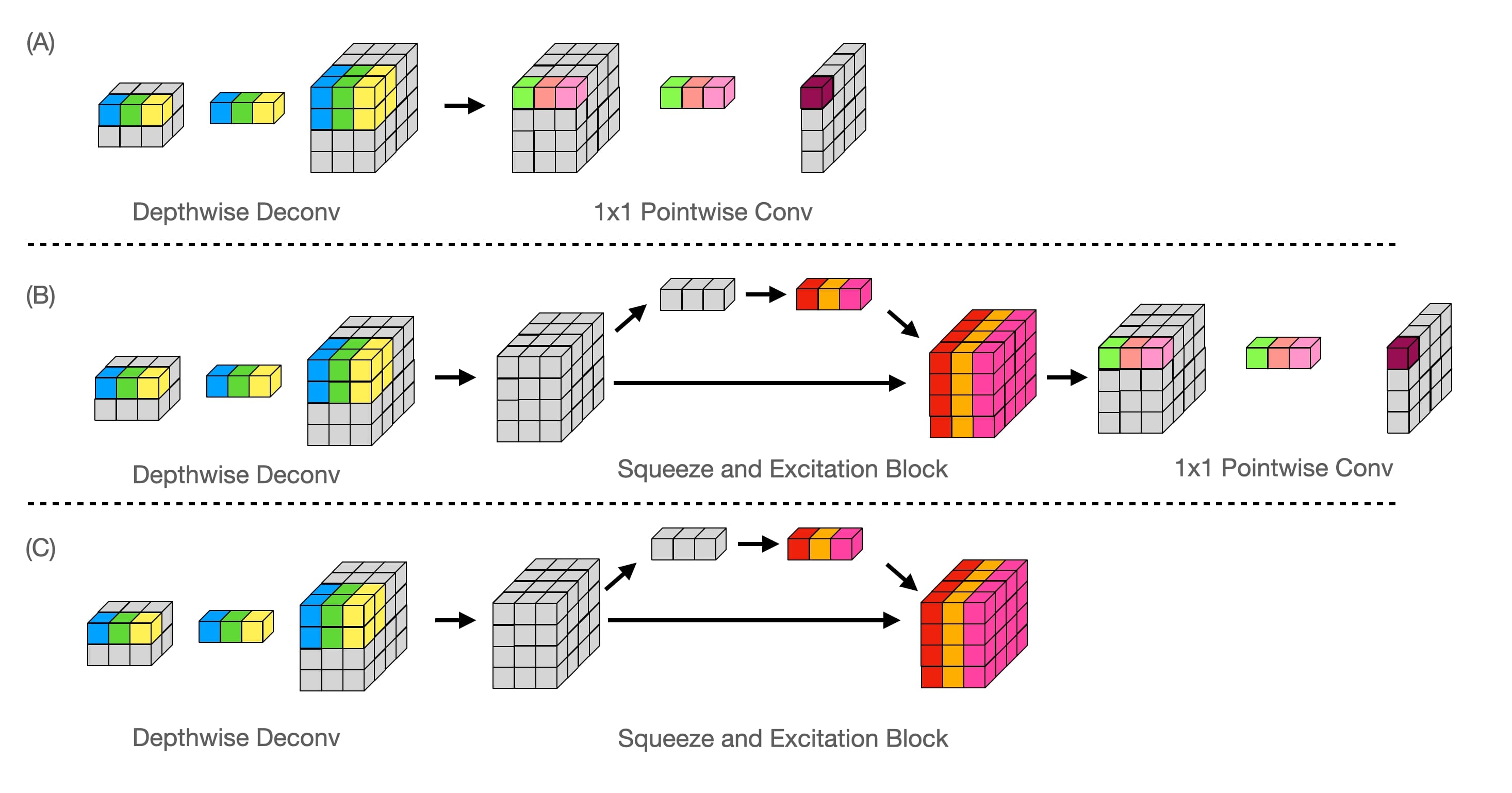}
\end{center}
\caption{We compare 3 different architectures for the pose and mesh decoders and the heatmap aggregator module. (A) follows a depthwise separable convolution design proposed in MobileNet~V1. (B) adopts the MBConv design for upsampling, which adds a squeeze-and-excitation block between depthwise deconvolution and pointwise convolution in (A). (C) is inspired by the fact that pointwise convolution takes most of the computation in (A) and (B). Therefore, we propose to eliminate it from (B), leading to our final design.}
\label{fig:decoding_block}
\end{figure}

Given the computation complexities shown in the Tab. \ref{tab:i2l_flops}, we introduce a lite version of I2L Mesh Net by replacing the original modules with its efficient counterparts. We start with the pose and mesh backbone, which directly adopts the ResNet50 structure in the original I2L Mesh Net. We replace it with an efficient model recently developed: the EfficientNet. We then design our own decoder and aggregation modules, which bring accuracy gains compared to the original decoder structure while at the same time being much more efficient when working with features extracted from the EfficientNet.

EfficientNets are a series of models with a scale-friendly baseline architecture developed through neural architecture search. Its main building block, the mobile inverted bottleneck MBConv was first proposed in \cite{sandler2018mobilenetv2} and later adopted by EfficientNet as the basic building block. EfficientNet introduces a series of models with various accuracy and computational cost trade-offs based on a fixed baseline structure determined through NAS and a compound scaling policy, which jointly scales up the model depth, width, and input resolution instead of focusing on scaling up a single dimension. To replace ResNet50 used in the pose and mesh backbone, we adopted the EfficientNet-B0 structure, which has slightly better classification performance on ImageNet with only $10\%$ computations. 

Inspired by the depthwise separable convolution proposed in MobileNet~V1 and the MBConv used in both EfficientNet and MobileNet~V2, we compared three possible architectures shown in Fig. \ref{fig:decoding_block} for pose and mesh decoders, as well as heatmap aggregators to perform efficient upsampling. Different from original I2L Mesh Net, we use upsampling in the heatmap aggregator as well. Our first structure (Fig. \ref{fig:decoding_block}.(A)) uses depthwise deconvolutions to perform upsampling, followed by $1\times1$ pointwise convolutions to aggregate information along the channels. Our second structure (Fig. \ref{fig:decoding_block}.(B)) uses MBConv as its basic building block, which introduces a squeeze and excitation (SE) block (\cite{hu2018squeeze}) between the depthwise deconvolution and the pointwise convolutoin to reweigh each channel. Lastly, inspired by the observation that the pointwise convolution takes most computations in both depthwise-separable convolution and MBConv, we proposed the last structure shown in Fig. \ref{fig:decoding_block}.(C), which skips the pointwise convolution of MBConv. For all three architectures, when the number of channels is inconsistant (e.g. in the first upsampling layer in the decoder module and the heatmap aggregation module), we add a pointwise convolution layer to change the width (i.e., the number of channels) of the models. 
We experiment with all three variations and empirically show that the architecture shown in Fig. \ref{fig:decoding_block}(C) provides the best accuracy-complexity trade-offs compared to the other designs in our experiments section.

\subsubsection{Generating Synthetic Data for Data Augmentation}
\label{sec:syn_data}

\begin{figure}[h!]
\begin{center}
\includegraphics[width=\linewidth]{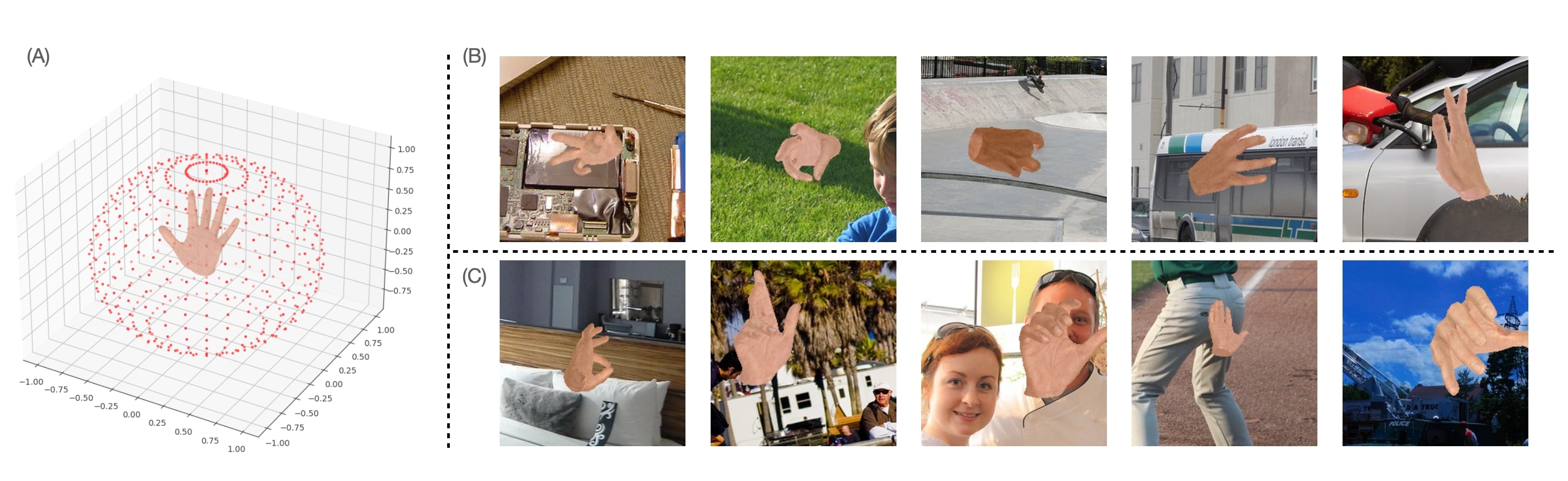}
\end{center}
\caption{We augment the training dataset by adding synthetic rendering data to improve the pose diversity and the camera view angle diversity. We visualize the example synthetic images with various pose in (B) and samples with diverse camera angles in (C). The camera locations that we utilized to create (C) are visualized in (A).}
\label{fig:syn_data}
\end{figure}

Although the EfficientNet-B0 obtains slightly higher accuracy compared to the ResNet50 on ImageNet, it may not generalize well to the hand pose estimation task we have and need to be refined with a significant amount of training data for hand pose estimation. We introduce an automatic pipeline to generate synthetic data without human intervention. 

We adopt two directions to further increase the diversity of the existing dataset following \cite{chen2022mobrecon}: 1) improving the pose diversity and 2) improving the view angle diversity. To boost the pose diversity, we leverage the 895 fitting poses shipped along with the MANO hand mesh. Furthermore, following \cite{zhou2020monocular}, we assume the poses of each finger is independent of each other, and further augment the 895 fitting poses by random swapping of finger poses. This design greatly increases the pose diversity of the hands. We denote the synthetic data constructed this way as \textbf{ManoRender}, which contains 57280 samples. To augment the camera view angles, we sample the camera positions on the unit sphere around the hand mesh. We fix the global rotation of the hand mesh to be upright and sample the camera positions on the unit sphere with an elevation of $[-\frac{\pi}{3}, \frac{\pi}{2}]$ using a step size of $\frac{\pi}{36}$. For azimuth, we sample in the range of $[0, 2\pi]$ with a step size of $\frac{\pi}{36}$. The camera locations are visualized in Fig. \ref{fig:syn_data} A. We denote the resulting dataset as \textbf{AngRender}, which contains 32 random samples at each camera position, totaling 71424 samples.

Finally, to render realistic hands, we adopt the hand appearance model HTML~\cite{qian2020html}, which provides a diverse set of hand skin textures for human hands. The HTML is a textured parametric hand model with textures based on the principal component analysis (PCA). It comes with an average appearance texture map $\bar{A}$ and 101 texture offset maps $A_\mathrm{offset}^i$. The final texture map could be represented as $A_\mathrm{final} = \bar{A}+\sum_{i=1}^{101}\alpha_i A_\mathrm{offset}^i$. We randomly sample $\alpha_i$ from a normal distribution $\mathcal{N}(0,\,1)$ to generate a random variety of hand appearances. For backgrounds, we randomly sample images from the MSCOCO dataset \cite{lin2014microsoft} and then randomly crop them to the input resolution. The example rendering results are visualized in Fig. \ref{fig:syn_data} (B) and (C). As shown in Fig. \ref{fig:syn_data}, the rendered images present a variety of hand poses and view angles, and also look realistic, thanks to the additional synthetic datasets we introduced and the high-quality textures provided by HTML. 

\subsection{Biomechanically Feasible IK Net}

\begin{figure}[h!]
\begin{center}
\includegraphics[width=\linewidth]{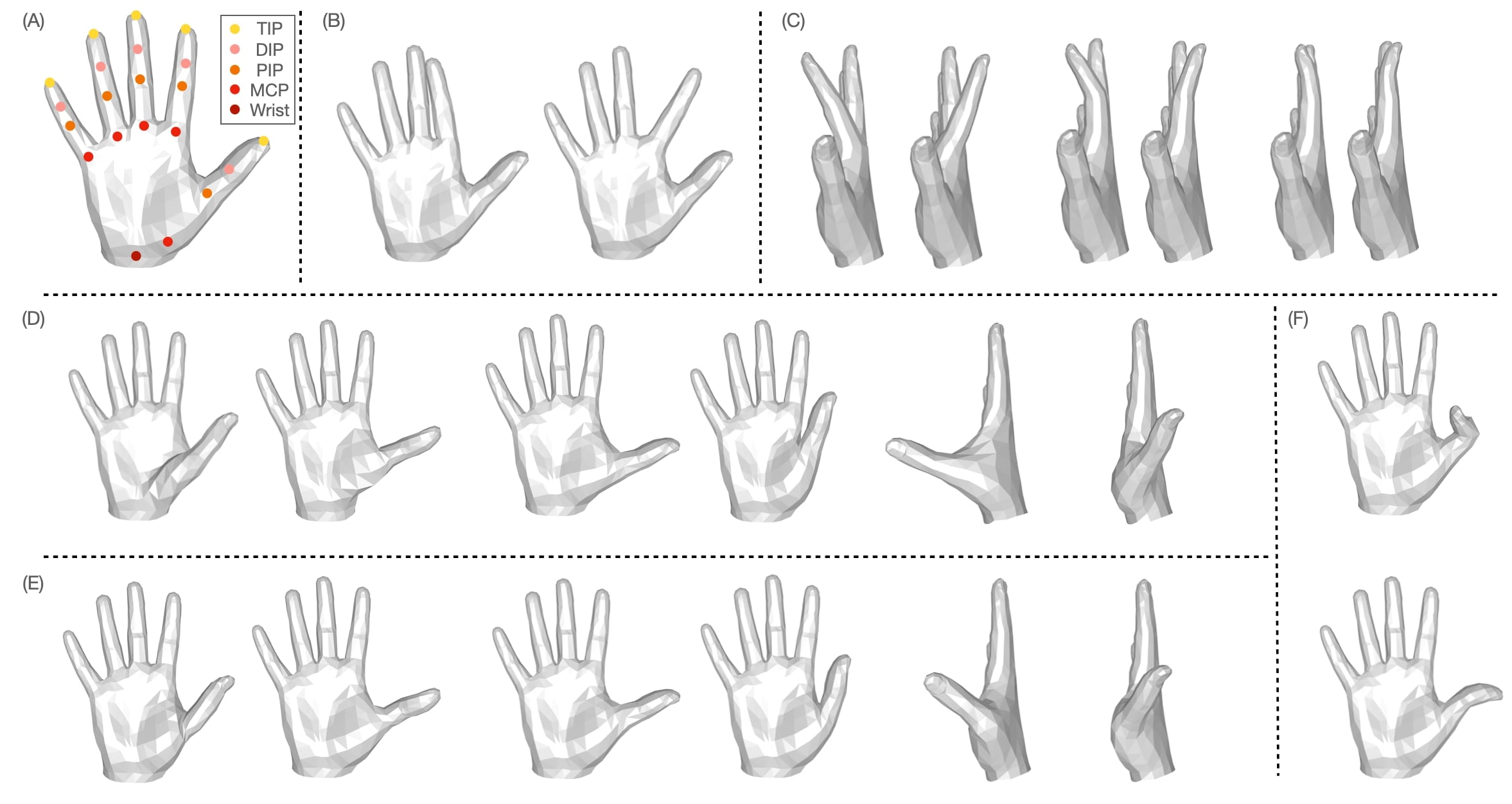}
\end{center}
\caption{We visualize the hand mesh to illustrate the restricted degree of freedom that our biomechanically feasible IK Net considers. (A) The original hand mesh in rest (neutral) pose and the definition of the joints on the hand. (B) The adduction and abduction movement of the index finger MCP joint. Similar movements are allowed for the middle finger, ring finger and little finger. (C) The Flexion and Extension movement of the MCP, PIP and DIP joint (from left to right). Similar movements are allowed for the middle finger, ring finger and little finger. (D) The rotation, adduction, abduction, flexion, and extension movement of the MCP joint of the thumb. (E) Similar set of movements allowed for the PIP joint of the thumb. (F) The Flexion and Extension movement of the DIP joint of the Thumb. }
\label{fig:hand_dof}
\end{figure}

The lite I2L Mesh Net directly predicts the 3D root-relative joint coordinates of the mesh without going through the rotation parameters of the parametric hand mesh (MANO). However, in many practical applications, rotations at each joint are required, for example, to control robotic hands to emulate human movement. In this section, we consider how to convert the mesh vertex positions to the rotation parameters. 

The original MANO hand mesh provides maximum flexibility for deforming the hand mesh and assumes 3 degrees of freedom for each of the joints. However, some of the rotations are infeasible for most of the human beings. For example, most people have trouble rotating the index finger along the roll axis. Furthermore, most robotic hands have a limited degree of freedom, making it necessary for us to limit the degree of freedom of hand movements if targeting robotic applications. We reduce the original 45 degree of freedom of the articulated MANO model to 23 degree of freedom following \cite{lim2020mobilehand}, which involves 4 DoF for each finger and 7 DoF for the thumb, as illustrated in Fig. \ref{fig:hand_dof}. We prepared the transformation between 45 DoF and 23 DoF based on a standard hand mesh and then used the transformation for all samples.

\begin{figure}[h!]
\begin{center}
\includegraphics[width=\linewidth]{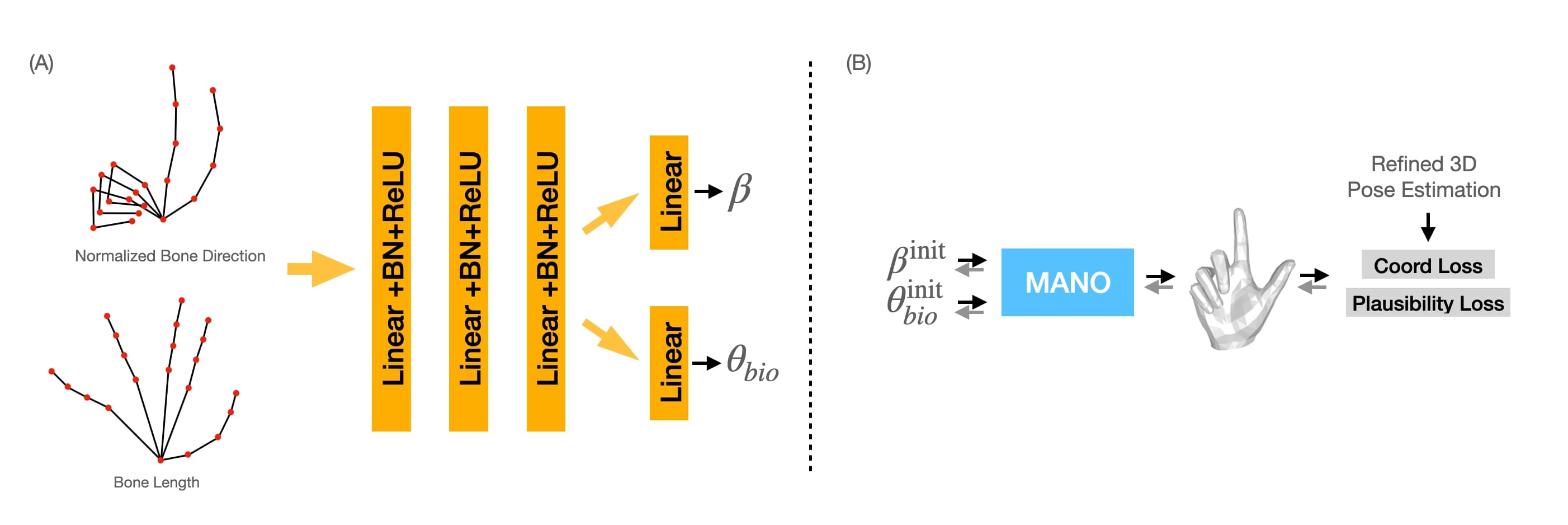}
\end{center}
\caption{(A) The proposed biomechanically feasible IK net. It takes the normalized bone direction and bone length as input and predicts the MANO shape parameters $\beta$ and the set of biomechanically feasible rotation parameters $\theta_{bio}$. (B) The optimization-based post-processing module. It iteratively updates the initial estimated shape parameters $\beta^{init}$ and the rotation parameters $\theta_{bio}^{init}$ to minimize the fitting error with respect to the predicted 3D joint and mesh coordinates by Lite I2L MeshNet. An additional regularization term is used to penalize unnatural poses like impossible bending directions.}
\label{fig:IK}
\end{figure}

We introduce a neural net model to infer the restricted but biomechanically feasible set of joint rotation parameters (also known as pose or kinematic parameters) of the MANO model from the estimated joint positions following \cite{zhou2020monocular, moon2020i2l}. The process of deriving the rotations at the joints based on the final locations of the joints is called inverse kinematics (IK). In the pose estimation literature, there are usually two directions to address this problem: 1) the optimization-based approach and 2) the learning-based approach. We opt for the learning-based method since previous work \cite{zhou2020monocular} shows that the optimization-based inverse kinematic process from scratch is often slow and could easily be trapped in local minima due to the high non-linearity of the IK process. 
Our IK net consists of three fully connected layers, each of which contains a linear operation followed by batch normalization and ReLU. Two additional heads are introduced to regress the 10 shape parameters and the 23 restricted rotation parameters separately. 
Instead of directly feeding the estimated 3D joint coordinates as in \cite{moon2020i2l}, inspired by \cite{zhou2020monocular}, we prepare the input by converting the raw coordinates into normalized direction and bone length, which is more explicit to infer the articulation of the skeleton and the shape of the hand. Different from \cite{zhou2020monocular}, we omit the reference bone length and bone direction, as we found empirically it doesn't provide gains. The proposed IK net architecture is shown in Fig. \ref{fig:IK} (A).

\subsection{Postprocessing through Optimization}

Due to the limited amount of training data and the restricted set of rotation parameters $\theta_{bio}$, the reconstructed mesh using the shape parameters $\beta$ and rotation parameters $\theta$ estimated by the IK net is often slightly misaligned with the mesh initially estimated using the lite I2L mesh net, which is often more accurate and reliable. To refine the initial estimated $\beta_{init}$ and $\theta_{bio}^{init}$, inspired by the optimization in the loop idea \cite{kolotouros2019learning, lv2021handtailor}, we introduce an optional post-optimization module, as illustrated in Fig. \ref{fig:IK} (B). The post-optimization module minimizes the fitting error with respect to the initially estimated 3D joint and mesh vertex positions. It also imposes an additional loss to penalize unfeasible poses. 

Specifically, we treat the 3D joint and mesh coordinates estimated by the Lite I2L mesh net as the pseudo-ground truth and minimize the fitting error by updating the shape and rotation parameters. To further penalize unnatural poses, we adopt an additional regularization term to penalize unnatural finger poses following \cite{zhang2019end, lv2021handtailor}:
\begin{equation}
    (\vec{B}_\mathrm{TIP-DIP}\times\vec{B}_\mathrm{DIP-PIP})\cdot(\vec{B}_\mathrm{DIP-PIP}\times\vec{B}_\mathrm{PIP-MCP}) \ge 0
\end{equation}
where $\vec{B}_\mathrm{a-b}$ indicates the bone vector from joint $J_b$ to $J_a$. We apply this geometric constraint to the index finger, middle finger, ring finger, and little finger. Intuitively, this loss penalizes the different bend direction at the PIP and the DIP joint. Note that, thanks to our bioinspired IK net design, the unnatural twisted finger pose will not occur in our predictions. Therefore, we do not need to introduce additional penalization in \cite{zhang2019end, lv2021handtailor} to penalize the twist rotations along the finger. 

\section{Experiments}

\subsection{Implementation Details}

We train the network using Adam Optimizer \cite{kingma2014adam} with a batch size of 32. For the Lite I2L mesh net, we train the networks for 30 epochs, with an initial learning rate of 1e-4. The learning rate decays to the $1/10$th in the $28$th epoch. The input resolution is $256\times256$. We follow the loss of the original I2L Mesh Net. For the lixel heatmap prediction, the loss is defined as:

\begin{equation}
    L = L_\mathrm{pose}^\mathrm{PoseNet}+L_\mathrm{pose}^\mathrm{MeshNet}+L_\mathrm{vertex}+L_\mathrm{edge}+0.1\times L_\mathrm{normal}
\end{equation}

Please refer to the original I2L Mesh Net \cite{moon2020i2l} for the details of each loss term (i.e., $L_\mathrm{pose}^\mathrm{PoseNet}$, $L_\mathrm{pose}^\mathrm{MeshNet}$, $L_\mathrm{vertex}$, $L_\mathrm{edge}$ and $L_\mathrm{normal}$). For training the Bio-inspired IK net, we run 40 epochs with learning rate decreased to 1/10 at the 30th and 35th epoch. We use the $L1$ loss on the shape and the restricted pose parameters as well as $L1$ loss on the reconstructed joint coordinates following:
\begin{align}
    L &= L_\theta^{bio}+L_\beta+L_\mathrm{pose}\\
    L_\theta^{bio} & = |\theta^{pred}_{bio}-\theta^{gt}_{bio}|\\
    L_\beta &= |\beta^{pred}-\beta^{gt}|\\
    L_\mathrm{pose} &= |\mathcal{J}(\mathcal{M}(\theta_{bio}, \beta))-P_{gt}|
\end{align}
where $\theta^{pred}_{bio}$ and $\beta^{pred}$ are the predicted restricted pose parameters and the shape parameters, $\theta^{gt}_{bio}$ and $\beta^{gt}$ are accordingly the ground truth pose and shape parameters. $\mathcal{M}(\theta_{bio}, \beta)$ is the reconstructed mesh based on the $\theta_{bio}$ and the $\beta$ and $\mathcal{J}$ is the joint regressor, which regresses the joint position based on the positions of the mesh vertices. $P_{gt}$ is the ground truth position of the hand skeletons.

\subsection{Dataset and Metrics}
Following the original I2L Mesh Net, we train and evaluate our lite I2L Mesh Net using the FreiHand dataset, which consists of 130,240 training images and
3,960 evaluation samples. Of the 130,240 training images, 32,560 images are original, and the remaining images are augmented from the original images by changing the background. The ground truth of the evaluation samples is not available, so we submit the results to the competition website for evaluation. We evaluate the networks using several commonly used metrics, which are detailed below.

\textbf{MPJPE/MPVPE}: evaluates the mean error per joint/vertex position by Euclidean distance (mm) between the estimated and ground-truth coordinates, respectively.

\textbf{PA-MPJPE/MPVPE}: first perform a procrustes analysis \cite{} to align estimated and ground-truth poses and then evaluate MPJPE / MPVPE.

\textbf{F-Score}: is the harmonic mean between the recall and the precision of two sets of points evaluated at a specific distance threshold (\cite{Knapitsch2017}). For example, "F@5mm" corresponds to a distance threshold of 5mm.


\subsection{Experimental Results and Discussion}

\begin{table}[t]
\centering
\small
\caption {Effect of the different model architectures for the backbone, the decoder and heatmap aggregation module in the I2L Mesh Net on the original FreiHand dataset. The best number is highlighted in \textbf{bold}, and the second best number is highlighted with an \underline{underline}. The baseline corresponds to the original I2L Mesh Net, which uses ResNet50 for pose/mesh backbones and conventional (de)convolution layers for the decoder and heatmap aggregation module. Ef-B0 Bb denotes the model with EfficientNet-B0 replacing the original ResNet50 as the backbone. The decoder remains the original design if not mentioned specifically. Dw Dec., MBConv Dec. and Dw SE Dec. denotes the decoder design using depthwise deconvolution (Fig. \ref{fig:decoding_block}(A)), MBConv based deconvolution (Fig. \ref{fig:decoding_block}(B)) and depthwise deconvolution with squeeze and excitation blocks for deconvolution (\ref{fig:decoding_block}(C)), respectively. Please refer to Sec. \ref{sec:ablation} for more details.} 
\begin{tabular}{lccccccc}
\label{tab:ablation_net}
\\\hline
Modules       & PA-MPJPE & PA-MPVPE &  F@5  &  F@15  &  GMacs  & \#Param.\\
\hline
Baseline                & \textbf{7.4} & \textbf{7.6} & \textbf{0.681} &  \textbf{0.973} &  29.06  &  136.63M\\
Ef-B0 Bb.               & 8.0 & 8.1 & 0.649 &  0.969 &  15.66  &  66.65M \\
Ef-B0 Bb. + Dw Dec.     & 8.1 & 8.2 & 0.642 &  0.967 &  \underline{5.43}   &  \underline{51.71M} \\
Ef-B0 Bb. + MBConv Dec. & 7.9 & 8.0 & 0.655 &  0.969 &  5.53   &  52.40M \\
Ef-B0 Bb. + Dw SE Dec.  & \underline{7.6} & \underline{7.7} & \underline{0.670} &  \textbf{0.973} &  \textbf{3.75}   &  \textbf{51.49M} \\
\hline
\end{tabular}
\end{table}

\subsubsection{Ablation Studies}
\label{sec:ablation}
\textbf{Network Architectures:} We first evaluate the accuracy and computation complexity trade-offs of using different backbone and decoder architectures in the Lite I2L Mesh Net. We start with the original I2L Mesh Net, which uses ResNet50 for the pose/mesh backbone and ordinary deconvolution layers for the decoder as well as the heatmap aggregator module. This model corresponds to the baseline in Tab. \ref{tab:ablation_net}. 

Our first variation focuses on improving the efficiency of the pose/mesh backbone. We replace the original ResNet50 with EfficientNet B0 as the backbone for feature extraction in both the PoseNet and the MeshNet. The remaining modules are left untouched. This model is denoted as "EF-B0 Bb." in Tab. \ref{tab:ablation_net}.

We then experiment with various decoder and heatmap aggregator designs. The "Dw Dec." denotes the decoder using depthwise separable deconvolutions as its basic building block (Fig.\ref{fig:decoding_block} (A)). 
We also experiment with an MBConv based decoder, denoted as "MBConv Dec." (Fig.\ref{fig:decoding_block} (B)). 
Finally, we experiment with our proposed decoder using the building block consisting of one depthwise deconvoluton layer and one squeeze and excitation block. 
This design is denoted as "Dw SE Deconv." in the tab. \ref{tab:ablation_net}.

As shown in the Tab. \ref{tab:ablation_net}, replacing the original ResNet50-based feature extractor with EfficientNet-B0 reduces the computational complexity by half. But the accuracy also drops.  When we use depthwise separable deconvolution as the basic building block for the decoders, the performance further drops, but the complexity also drops to around $18\%$ of the baseline. When we adopt MBConv-style building blocks, we were able to recover part of the lost accuracy. Finally, our proposed depthwise deconvolution and SE block based decoder architecture achieves the best accuracy and efficiency trade-off, thanks to the effective squeeze and excitation block, with only $13\%$ of computation complexity, while maintaining an accuracy very close to the baseline. 

\textbf{Synthetic Data Augmentation} We then investigate the performance gain induced by the introduction of synthetic data for model training. As mentioned in Sec. \ref{sec:syn_data}, we consider two different approaches to construct the synthetic data, one increases the pose diversity, while the other increases the camera view diversity. We denote the synthetic data created by each approach by ManoRender and AngRender, respectively. In the Tab. \ref{tab:abaltion_data}, we use MR to refer to the ManoRender dataset and AR for the AngRender dataset. We present the improvements for both the original I2L Mesh Net and our lightweight versions in the Tab. \ref{tab:abaltion_data}.

Compared to baseline training on the FreiHand dataset alone, the introduction of either ManoRender or AugRender dataset improves accuracy. The best performance was obtained when both synthetic datasets were introduced. When compared between models, both the proposed lite version of I2L Mesh Net and the original I2L Mesh Net benefit from the synthetic data set, indicating the efficacy of the proposed automated synthetic data generation pipeline. 

\begin{table}[t]
\centering
\small
\caption {Effect of synthetic data augmentation. We evaluate the performance gain by augmenting original FreiHand training data with synthetic data created for increasing different diversities. MR denotes the synthetic ManoRender dataset, which improves the pose diversity while AR denotes the synthetic AngRender dataset, which improves the camera view diversity. Frei is the original FreiHand dataset. Performance reported for the Frei evaluation dataset.} 
\begin{tabular}{lcccccccc}
\label{tab:abaltion_data}
\\\hline
Model.   &  Dataset             & PA-MPJPE & PA-MPVPE &  F@5  &  F@15  & GMacs \\\hline
Original I2L Mesh Net &  Frei                &  7.4     &   7.6    & 0.681 &  0.943 & 29.06\\
Original I2L Mesh Net &  Frei+MR             &  6.9     &   7.0    & 0.707 &  0.980 & 29.06 \\
Original I2L Mesh Net &  Frei+AR             &  6.9     &   7.0    & 0.709 &  0.980 & 29.06 \\
Original I2L Mesh Net &  Frei+MR+AR          &  6.7     &   6.8    & 0.719 &  0.981 & 29.06 \\\hline
Lite I2L Mesh Net & Frei     &  7.6     &   7.7    & 0.670 &  0.973 & 3.75 \\
Lite I2L Mesh Net & Frei+MR  &  7.4     &   7.5    & 0.680 &  0.975 & 3.75 \\
Lite I2L Mesh Net & Frei+AR  &  7.3     &   7.4    & 0.687 &  0.975 & 3.75 \\
Lite I2L Mesh Net & Frei+MR+AR & 7.0    &   7.1    & 0.700 &  0.979 & 3.75 \\
\hline
\end{tabular}
\end{table}

\textbf{Biomechanically feasible IK Net}. Finally, we investigate the benefit from using the proposed IK net, as well as the optional post-optimization module in Tab.~\ref{tab:ablation_ik}. All results are obtained by using our Lite I2L Mesh Net to generate the initial mesh, and then applying an IK net to generate the rotation and shape parameters with optional post-optimization. The baseline IK net regresses to 45 rotation parameters, while the biomechanically feasible IK net regresses to 23 rotation parameters.   Compared to using 45 rotation parameters,  restricting the DoF to 23 slightly lowers the accuracy in terms of joint and mesh vertex error. This is because we impose stronger constraints on the mesh structure and, therefore, reduce the representation capacity of the parametric hand mesh. However, with only 20 iterations of mesh fitting using the optional post-optimization module, the model could not only recover the lost accuracy, but also surpass the baseline. This shows the effectiveness of the post-optimization module.

In certain practical applications, the shape of the hand in the natural pose could be obtained in advance by 3D scan. This is the case, for example, for telerobotic surgery or rehabilitation, where the robotic hand is intended to mimic the hand of a particular surgeon or therapist.  In such scenarios, we could compute the shape parameters $\beta$ before hand. Therefore, we also provide results for the IK net with known shape parameters. However, since the ground truth shape parameters are not provided for the FreiHand evaluation set, we split the training set into a train-val subset and a validation subset with a ratio of $9:1$. Specifically, we perform the split based on the original images and then extend to the full augmented set (i.e., the augmented images from an original image are put into the same subset as the original image) to avoid data leakage. We denote experiments that perform training on the train-val set and evaluate on the validation set with $\dagger$. Tab. \ref{tab:ablation_ik} shows that, when the user-specific shape information can be predetermined, we can substantially improve the joint and mesh estimation accuracy. As in the case where the shape information is not known, post-optimization can significantly improve the accuracy beyond using the IK net only.

\begin{table}[t]
\centering
\small
\caption {Effect of the biomechanically feasible IK Net and the post-optimization module. Experiments marked with $\dagger$ are performed with known shape information and are evaluated on a validation set splitted from the training set because the ground truth shape parameters $\beta$ are not available for the evaluation set.} 
\begin{tabular}{lcccccccc}
\label{tab:ablation_ik}
\\\hline
Modules                                    & \#DoF & GT Shape & PA-MPJPE & PA-MPVPE &  F@5  &  F@15  \\\hline
Baseline IK  & 45+10 & \xmark & 7.1 & 7.3 & 0.686 & 0.979 \\
Baseline IK$^\dagger$   & 45 & $\cmark$ & 4.8 & 5.1 & 0.817 & 0.918 \\\hline
Our IK                          & 23+10  & \xmark & 7.4 & 7.6 & 0.668 & 0.978 \\
Our IK + Post Opt.              & 23+10 & \xmark & 6.9 & 7.0 & 0.706 & 0.980 \\\hline
Our IK$^\dagger$             & 23   & \cmark  & 4.8 & 5.0 & 0.831 & 0.998 \\
Our IK + Post Opt.$^\dagger$ & 23   & \cmark & 3.7 & 3.8 & 0.919 & 0.999 \\
\hline
\end{tabular}
\end{table}

\subsubsection{Comparison with State-Of-The-Art Hand Pose Estimation Methods }

\begin{table}[t]
\centering
\small
\caption {Comparison with the state-of-the-art models. We abbreviate PA-MPJPE and PA-MPVPE as PJ and PV. * denotes model with stacked architecture. OM/YT indicates using ObMan \cite{hasson19_obman}/Youtube 3D \cite{Kulon_2020_CVPR} as additional training data. Synthetic indicates using additional synthetic data for training. TTA corresponds to test time augmentation. Note that the computation cost of our entire lite I2L Mesh Net is less than a single ResNet-50 or HRNet backbone. Our Lite I2L MeshNet obtains state-of-the-art accuracy and efficiency trade-offs comparing to existing models.} 
\begin{tabular}{lccccccccc}
\label{tab:sota}
\\\hline
Model.                                 &  Backbone  & PJ & PV &  F@5  &  F@15  & Comment\\\hline
MobileHand \cite{lim2020mobilehand}    &  MobileNet &  -       &  13.1    & 0.439 &  0.902 \\
FreiHand \cite{zimmermann2019freihand} &  ResNet50  &  11.0    &  10.9    & 0.516 &  0.934 \\
YoutubeHand \cite{kulon2020weakly}     &  ResNet50  &  8.4     &   8.6    & 0.614 &  0.966 \\
Pose2Mesh \cite{Choi_2020_ECCV_Pose2Mesh} &  HRNet  &  7.4     &   7.6    & 0.683 &  0.973 \\
I2L-MeshNet \cite{moon2020i2l}         &  ResNet50*  &  7.4     &   7.6    & 0.681 &  0.973 \\
HIU-DMTL \cite{zhang2021hand}          & Customized* &  7.1     &   7.3    & 0.699 &  0.974 \\
CMR \cite{chen2021camera}              &  ResNet50*  &  6.9     &   7.0    & 0.715 &  0.977 \\
I2UV-HandNet \cite{chen2021i2uv}       &  ResNet50  &  6.7     &   6.9    & 0.707 &  0.977 & 150K OM+50k YT \\
METRO \cite{lin2021end}                &  HRNet     &  6.7     &   6.8    & 0.717 &  0.981 & TTA\\
Tang et al. \cite{tang2021towards}     &  ResNet50  &  6.7     &   6.7    & 0.724 &  0.981 & 150K OM\\
MeshGraphormer \cite{lin2021mesh}      &  HRNet     &  5.9     &   6.0    & 0.765 &  0.987 & TTA\\
MobRecon \cite{chen2022mobrecon}       &  ResNet50* &  5.7     &   5.8    & 0.784 &  0.986 & 328K Synthetic \\\hline
Lite I2L MeshNet (Frei)                             & EfficientNet-B0* &  7.6     &   7.7    & 0.670 &  0.973 & \\
Lite I2L MeshNet (Frei+MR+AR)                      & EfficientNet-B0* &  7.0     &   7.1    & 0.700 &  0.979 & 128K Synthetic\\\hline
\begin{tabular}{@{}c@{}} Lite I2L MeshNet (Frei+MR+AR) \\ + IK + Post Opt.\end{tabular}      &
EfficientNet-B0* &  6.9     &   7.0    & 0.706 &  0.980 & 128K Synthetic\\
\hline
\end{tabular}
\end{table}

Finally, we compare our models with the state-of-the-art models in the literature and present the results in Tab. \ref{tab:sota}. In addition to the evaluation metrics, we also provide additional information such as the backbone structure and whether to use additional data for training or test-time augmentation (TTA), which helps us better understand the performance of the models.

As shown in the Tab. \ref{tab:sota}, with the additional synthetic data, our lite I2L Mesh Net could obtain state-of-the-art performance and trade-off between accuracy and efficiency. Notice that the overall computational complexity of our model is only 3.75 GMacs, only 70\% of the ResNet50 backbone (5.38 GMacs)  and is $13\%$ of the total complexity of the original I2L Mesh Net. It is also substantially less complex than other state-of-the-art models including \cite{chen2021i2uv, tang2021towards, lin2021end, lin2021mesh, chen2022mobrecon}. This comparison demonstrates both the efficacy and the efficiency of our proposed light weight I2L Mesh Net for joint and mesh estimation. Adding our IK and post-optimization process can slightly improve the estimation accuracy at only a small computational cost. The IK and post-optimization modules can be applied to other joint and mesh estimation models as well for applications requiring the translation of the hand mesh to rotation parameters.  

\section{Conclusion}

We present a lite version of the state-of-the-art I2L Mesh Net to estimate the position of the hand joints and the hand mesh. We approach this problem by first profiling the original I2L Mesh Net and then introducing efficient modules to replace the corresponding modules in the original model. With the EfficientNet backbone and the proposed lightweight yet effective basic building blocks for the decoder and the heatmap aggregator, the proposed lightweight model can reduce the computational complexity by a factor of 7.6, while achieving an estimation accuracy for hand joints and mesh vertices very close to the original model.
We also introduce an automatic pipeline to generate synthetic data to enhance the pose and viewing angle diversity of the training data, which is shown to significantly improve the estimation accuracy on the testing data.  
Lastly, we also introduce a lightweight inverse kinematic (IK) module to predict a biomechanically feasible set of joint rotation parameters and shape parameters from the mesh vertices derived by the Lite I2L Mesh Net. The predicted rotation and shape parameters could be further fed into an optional post-optimization module so that the reconstructed mesh from the final rotation and shape parameters is better aligned with the initial estimated mesh. Such optimization leads to more accurate parameter estimations and more accurate mesh reconstructions. 
We conduct extensive experiments with our proposed model and compare its performance to the state-of-the-art.  Our Lite I2L Mesh Net trained with augmented synthetic data obtains comparable estimation accuracy with the state-of-the-art models at a fraction of the computational cost. Combining the lite I2L Mesh Net with the biomechanically feasible IK net and post-optimization can further improve the accuracy slightly, but more importantly, provide the kinematic parameters that are important for applications that leverage hand pose estimation to control robotic hands. We further show that in applications where user-specific hand shape can be measured in advance, the estimation accuracy can be greatly improved.

\newpage
\bibliographystyle{Frontiers-Harvard} 
\bibliography{test}

\begin{thebibliography}{60}
\providecommand{\natexlab}[1]{#1}
\expandafter\ifx\csname urlstyle\endcsname\relax
  \providecommand{\doi}[1]{doi:\discretionary{}{}{}#1}\else
  \providecommand{\doi}{doi:\discretionary{}{}{}\begingroup
  \urlstyle{rm}\Url}\fi
\providecommand{\selectlanguage}[1]{\relax}
\providecommand{\bibAnnoteFile}[1]{%
  \IfFileExists{#1}{\begin{quotation}\noindent\textsc{Key:} #1\\
  \textsc{Annotation:}\ \input{#1}\end{quotation}}{}}
\providecommand{\bibAnnote}[2]{%
  \begin{quotation}\noindent\textsc{Key:} #1\\
  \textsc{Annotation:}\ #2\end{quotation}}

\bibitem[{Boukhayma et~al.(2019)Boukhayma, Bem, and Torr}]{boukhayma20193d}
Boukhayma, A., Bem, R.~d., and Torr, P.~H. (2019).
\newblock 3d hand shape and pose from images in the wild.
\newblock In \emph{Proceedings of the IEEE/CVF Conference on Computer Vision
  and Pattern Recognition}. 10843--10852
\bibAnnoteFile{boukhayma20193d}

\bibitem[{Cai et~al.(2018)Cai, Ge, Cai, and Yuan}]{cai2018weakly}
Cai, Y., Ge, L., Cai, J., and Yuan, J. (2018).
\newblock Weakly-supervised 3d hand pose estimation from monocular rgb images.
\newblock In \emph{Proceedings of the European Conference on Computer Vision
  (ECCV)}. 666--682
\bibAnnoteFile{cai2018weakly}

\bibitem[{Chen et~al.(2021{\natexlab{a}})Chen, Chen, Yang, Wu, Li, Xia
  et~al.}]{chen2021i2uv}
Chen, P., Chen, Y., Yang, D., Wu, F., Li, Q., Xia, Q., et~al.
  (2021{\natexlab{a}}).
\newblock I2uv-handnet: Image-to-uv prediction network for accurate and
  high-fidelity 3d hand mesh modeling.
\newblock In \emph{Proceedings of the IEEE/CVF International Conference on
  Computer Vision}. 12929--12938
\bibAnnoteFile{chen2021i2uv}

\bibitem[{Chen et~al.(2022)Chen, Liu, Dong, Zhang, Ma, Xiong
  et~al.}]{chen2022mobrecon}
Chen, X., Liu, Y., Dong, Y., Zhang, X., Ma, C., Xiong, Y., et~al. (2022).
\newblock Mobrecon: Mobile-friendly hand mesh reconstruction from monocular
  image.
\newblock In \emph{Proceedings of the IEEE/CVF Conference on Computer Vision
  and Pattern Recognition}. 20544--20554
\bibAnnoteFile{chen2022mobrecon}

\bibitem[{Chen et~al.(2021{\natexlab{b}})Chen, Liu, Ma, Chang, Wang, Chen
  et~al.}]{chen2021camera}
Chen, X., Liu, Y., Ma, C., Chang, J., Wang, H., Chen, T., et~al.
  (2021{\natexlab{b}}).
\newblock Camera-space hand mesh recovery via semantic aggregation and adaptive
  2d-1d registration.
\newblock In \emph{Proceedings of the IEEE/CVF Conference on Computer Vision
  and Pattern Recognition}. 13274--13283
\bibAnnoteFile{chen2021camera}

\bibitem[{Chen et~al.(2021{\natexlab{c}})Chen, Tu, Kang, Bao, Zhang, Zhe
  et~al.}]{chen2021model}
Chen, Y., Tu, Z., Kang, D., Bao, L., Zhang, Y., Zhe, X., et~al.
  (2021{\natexlab{c}}).
\newblock Model-based 3d hand reconstruction via self-supervised learning.
\newblock In \emph{Proceedings of the IEEE/CVF Conference on Computer Vision
  and Pattern Recognition}. 10451--10460
\bibAnnoteFile{chen2021model}

\bibitem[{Chessa et~al.(2019)Chessa, Maiello, Klein, Paulun, and
  Solari}]{chessa2019grasping}
Chessa, M., Maiello, G., Klein, L.~K., Paulun, V.~C., and Solari, F. (2019).
\newblock Grasping objects in immersive virtual reality.
\newblock In \emph{2019 IEEE Conference on Virtual Reality and 3D User
  Interfaces (VR)} (IEEE), 1749--1754
\bibAnnoteFile{chessa2019grasping}

\bibitem[{Choi et~al.(2020)Choi, Moon, and Lee}]{Choi_2020_ECCV_Pose2Mesh}
Choi, H., Moon, G., and Lee, K.~M. (2020).
\newblock Pose2mesh: Graph convolutional network for 3d human pose and mesh
  recovery from a 2d human pose.
\newblock In \emph{European Conference on Computer Vision (ECCV)}
\bibAnnoteFile{Choi_2020_ECCV_Pose2Mesh}

\bibitem[{Ge et~al.(2018)Ge, Cai, Weng, and Yuan}]{ge2018hand}
Ge, L., Cai, Y., Weng, J., and Yuan, J. (2018).
\newblock Hand pointnet: 3d hand pose estimation using point sets.
\newblock In \emph{Proceedings of the IEEE Conference on Computer Vision and
  Pattern Recognition}. 8417--8426
\bibAnnoteFile{ge2018hand}

\bibitem[{Ge et~al.(2019)Ge, Ren, Li, Xue, Wang, Cai et~al.}]{ge20193d}
Ge, L., Ren, Z., Li, Y., Xue, Z., Wang, Y., Cai, J., et~al. (2019).
\newblock 3d hand shape and pose estimation from a single rgb image.
\newblock In \emph{Proceedings of the IEEE/CVF Conference on Computer Vision
  and Pattern Recognition}. 10833--10842
\bibAnnoteFile{ge20193d}

\bibitem[{Gholami et~al.(2018)Gholami, Kwon, Wu, Tai, Yue, Jin
  et~al.}]{gholami2018squeezenext}
Gholami, A., Kwon, K., Wu, B., Tai, Z., Yue, X., Jin, P., et~al. (2018).
\newblock Squeezenext: Hardware-aware neural network design.
\newblock In \emph{Proceedings of the IEEE Conference on Computer Vision and
  Pattern Recognition Workshops}. 1638--1647
\bibAnnoteFile{gholami2018squeezenext}

\bibitem[{Goodfellow et~al.(2014)Goodfellow, Pouget-Abadie, Mirza, Xu,
  Warde-Farley, Ozair et~al.}]{goodfellow2014generative}
Goodfellow, I., Pouget-Abadie, J., Mirza, M., Xu, B., Warde-Farley, D., Ozair,
  S., et~al. (2014).
\newblock Generative adversarial nets.
\newblock \emph{Advances in neural information processing systems} 27
\bibAnnoteFile{goodfellow2014generative}

\bibitem[{Hasson et~al.(2019{\natexlab{a}})Hasson, Varol, Tzionas, Kalevatykh,
  Black, Laptev et~al.}]{hasson2019learning}
Hasson, Y., Varol, G., Tzionas, D., Kalevatykh, I., Black, M.~J., Laptev, I.,
  et~al. (2019{\natexlab{a}}).
\newblock Learning joint reconstruction of hands and manipulated objects.
\newblock In \emph{Proceedings of the IEEE/CVF conference on computer vision
  and pattern recognition}. 11807--11816
\bibAnnoteFile{hasson2019learning}

\bibitem[{Hasson et~al.(2019{\natexlab{b}})Hasson, Varol, Tzionas, Kalevatykh,
  Black, Laptev et~al.}]{hasson19_obman}
Hasson, Y., Varol, G., Tzionas, D., Kalevatykh, I., Black, M.~J., Laptev, I.,
  et~al. (2019{\natexlab{b}}).
\newblock Learning joint reconstruction of hands and manipulated objects.
\newblock In \emph{CVPR}
\bibAnnoteFile{hasson19_obman}

\bibitem[{Howard et~al.(2019)Howard, Sandler, Chu, Chen, Chen, Tan
  et~al.}]{howard2019searching}
Howard, A., Sandler, M., Chu, G., Chen, L.-C., Chen, B., Tan, M., et~al.
  (2019).
\newblock Searching for mobilenetv3.
\newblock In \emph{Proceedings of the IEEE/CVF International Conference on
  Computer Vision}. 1314--1324
\bibAnnoteFile{howard2019searching}

\bibitem[{Howard et~al.(2017)Howard, Zhu, Chen, Kalenichenko, Wang, Weyand
  et~al.}]{howard2017mobilenets}
Howard, A.~G., Zhu, M., Chen, B., Kalenichenko, D., Wang, W., Weyand, T.,
  et~al. (2017).
\newblock Mobilenets: Efficient convolutional neural networks for mobile vision
  applications.
\newblock \emph{arXiv preprint arXiv:1704.04861}
\bibAnnoteFile{howard2017mobilenets}

\bibitem[{Hu et~al.(2018)Hu, Shen, and Sun}]{hu2018squeeze}
Hu, J., Shen, L., and Sun, G. (2018).
\newblock Squeeze-and-excitation networks.
\newblock In \emph{Proceedings of the IEEE conference on computer vision and
  pattern recognition}. 7132--7141
\bibAnnoteFile{hu2018squeeze}

\bibitem[{Iandola et~al.(2016)Iandola, Han, Moskewicz, Ashraf, Dally, and
  Keutzer}]{iandola2016squeezenet}
Iandola, F.~N., Han, S., Moskewicz, M.~W., Ashraf, K., Dally, W.~J., and
  Keutzer, K. (2016).
\newblock Squeezenet: Alexnet-level accuracy with 50x fewer parameters and< 0.5
  mb model size.
\newblock \emph{arXiv preprint arXiv:1602.07360}
\bibAnnoteFile{iandola2016squeezenet}

\bibitem[{Iqbal et~al.(2018)Iqbal, Molchanov, Gall, and Kautz}]{iqbal2018hand}
Iqbal, U., Molchanov, P., Gall, T. B.~J., and Kautz, J. (2018).
\newblock Hand pose estimation via latent 2.5 d heatmap regression.
\newblock In \emph{Proceedings of the European Conference on Computer Vision
  (ECCV)}. 118--134
\bibAnnoteFile{iqbal2018hand}

\bibitem[{Kingma and Ba(2014)}]{kingma2014adam}
Kingma, D.~P. and Ba, J. (2014).
\newblock Adam: A method for stochastic optimization.
\newblock \emph{arXiv preprint arXiv:1412.6980}
\bibAnnoteFile{kingma2014adam}

\bibitem[{Knapitsch et~al.(2017)Knapitsch, Park, Zhou, and
  Koltun}]{Knapitsch2017}
Knapitsch, A., Park, J., Zhou, Q.-Y., and Koltun, V. (2017).
\newblock Tanks and temples: Benchmarking large-scale scene reconstruction.
\newblock \emph{ACM Transactions on Graphics} 36
\bibAnnoteFile{Knapitsch2017}

\bibitem[{Kolotouros et~al.(2019)Kolotouros, Pavlakos, Black, and
  Daniilidis}]{kolotouros2019learning}
Kolotouros, N., Pavlakos, G., Black, M.~J., and Daniilidis, K. (2019).
\newblock Learning to reconstruct 3d human pose and shape via model-fitting in
  the loop.
\newblock In \emph{Proceedings of the IEEE/CVF International Conference on
  Computer Vision}. 2252--2261
\bibAnnoteFile{kolotouros2019learning}

\bibitem[{Kucuk and Bingul(2006)}]{kucuk2006robot}
Kucuk, S. and Bingul, Z. (2006).
\newblock \emph{Robot kinematics: Forward and inverse kinematics} (INTECH Open
  Access Publisher London, UK)
\bibAnnoteFile{kucuk2006robot}

\bibitem[{Kulon et~al.(2020{\natexlab{a}})Kulon, Guler, Kokkinos, Bronstein,
  and Zafeiriou}]{Kulon_2020_CVPR}
Kulon, D., Guler, R.~A., Kokkinos, I., Bronstein, M.~M., and Zafeiriou, S.
  (2020{\natexlab{a}}).
\newblock Weakly-supervised mesh-convolutional hand reconstruction in the wild.
\newblock In \emph{The IEEE/CVF Conference on Computer Vision and Pattern
  Recognition (CVPR)}
\bibAnnoteFile{Kulon_2020_CVPR}

\bibitem[{Kulon et~al.(2020{\natexlab{b}})Kulon, Guler, Kokkinos, Bronstein,
  and Zafeiriou}]{kulon2020weakly}
Kulon, D., Guler, R.~A., Kokkinos, I., Bronstein, M.~M., and Zafeiriou, S.
  (2020{\natexlab{b}}).
\newblock Weakly-supervised mesh-convolutional hand reconstruction in the wild.
\newblock In \emph{Proceedings of the IEEE/CVF conference on computer vision
  and pattern recognition}. 4990--5000
\bibAnnoteFile{kulon2020weakly}

\bibitem[{Lambeta et~al.(2021)Lambeta, Xu, Xu, Chou, Wang, Darrell
  et~al.}]{lambeta2021pytouch}
Lambeta, M., Xu, H., Xu, J., Chou, P.-W., Wang, S., Darrell, T., et~al. (2021).
\newblock Pytouch: A machine learning library for touch processing.
\newblock In \emph{2021 IEEE International Conference on Robotics and
  Automation (ICRA)} (IEEE), 13208--13214
\bibAnnoteFile{lambeta2021pytouch}

\bibitem[{Li et~al.(2021)Li, Bian, Zeng, Wang, Pang, Liu et~al.}]{li2021human}
Li, J., Bian, S., Zeng, A., Wang, C., Pang, B., Liu, W., et~al. (2021).
\newblock Human pose regression with residual log-likelihood estimation.
\newblock In \emph{ICCV}
\bibAnnoteFile{li2021human}

\bibitem[{Lim et~al.(2020)Lim, Jatesiktat, and Ang}]{lim2020mobilehand}
Lim, G.~M., Jatesiktat, P., and Ang, W.~T. (2020).
\newblock Mobilehand: Real-time 3d hand shape and pose estimation from color
  image.
\newblock In \emph{International Conference on Neural Information Processing}
  (Springer), 450--459
\bibAnnoteFile{lim2020mobilehand}

\bibitem[{Lin et~al.(2021{\natexlab{a}})Lin, Wang, and Liu}]{lin2021end}
Lin, K., Wang, L., and Liu, Z. (2021{\natexlab{a}}).
\newblock End-to-end human pose and mesh reconstruction with transformers.
\newblock In \emph{Proceedings of the IEEE/CVF Conference on Computer Vision
  and Pattern Recognition}. 1954--1963
\bibAnnoteFile{lin2021end}

\bibitem[{Lin et~al.(2021{\natexlab{b}})Lin, Wang, and Liu}]{lin2021mesh}
Lin, K., Wang, L., and Liu, Z. (2021{\natexlab{b}}).
\newblock Mesh graphormer.
\newblock In \emph{Proceedings of the IEEE/CVF International Conference on
  Computer Vision}. 12939--12948
\bibAnnoteFile{lin2021mesh}

\bibitem[{Lin et~al.(2014)Lin, Maire, Belongie, Hays, Perona, Ramanan
  et~al.}]{lin2014microsoft}
Lin, T.-Y., Maire, M., Belongie, S., Hays, J., Perona, P., Ramanan, D., et~al.
  (2014).
\newblock Microsoft coco: Common objects in context.
\newblock In \emph{European conference on computer vision} (Springer), 740--755
\bibAnnoteFile{lin2014microsoft}

\bibitem[{Liu et~al.(2021)Liu, Jiang, Xu, Liu, and Wang}]{liu2021semi}
Liu, S., Jiang, H., Xu, J., Liu, S., and Wang, X. (2021).
\newblock Semi-supervised 3d hand-object poses estimation with interactions in
  time.
\newblock In \emph{Proceedings of the IEEE/CVF Conference on Computer Vision
  and Pattern Recognition}. 14687--14697
\bibAnnoteFile{liu2021semi}

\bibitem[{Lu et~al.(2003)Lu, Metaxas, Samaras, and Oliensis}]{lu2003using}
Lu, S., Metaxas, D., Samaras, D., and Oliensis, J. (2003).
\newblock Using multiple cues for hand tracking and model refinement.
\newblock In \emph{2003 IEEE Computer Society Conference on Computer Vision and
  Pattern Recognition, 2003. Proceedings.} (IEEE), vol.~2, II--443
\bibAnnoteFile{lu2003using}

\bibitem[{Lv et~al.(2021)Lv, Xu, Yang, Qian, Mao, and Lu}]{lv2021handtailor}
Lv, J., Xu, W., Yang, L., Qian, S., Mao, C., and Lu, C. (2021).
\newblock Handtailor: Towards high-precision monocular 3d hand recovery.
\newblock \emph{arXiv preprint arXiv:2102.09244}
\bibAnnoteFile{lv2021handtailor}

\bibitem[{Ma et~al.(2018)Ma, Zhang, Zheng, and Sun}]{ma2018shufflenet}
Ma, N., Zhang, X., Zheng, H.-T., and Sun, J. (2018).
\newblock Shufflenet v2: Practical guidelines for efficient cnn architecture
  design.
\newblock In \emph{Proceedings of the European conference on computer vision
  (ECCV)}. 116--131
\bibAnnoteFile{ma2018shufflenet}

\bibitem[{Moon and Lee(2020)}]{moon2020i2l}
Moon, G. and Lee, K.~M. (2020).
\newblock I2l-meshnet: Image-to-lixel prediction network for accurate 3d human
  pose and mesh estimation from a single rgb image.
\newblock In \emph{European Conference on Computer Vision} (Springer), 752--768
\bibAnnoteFile{moon2020i2l}

\bibitem[{Mueller et~al.(2018)Mueller, Bernard, Sotnychenko, Mehta, Sridhar,
  Casas et~al.}]{mueller2018ganerated}
Mueller, F., Bernard, F., Sotnychenko, O., Mehta, D., Sridhar, S., Casas, D.,
  et~al. (2018).
\newblock Ganerated hands for real-time 3d hand tracking from monocular rgb.
\newblock In \emph{Proceedings of the IEEE Conference on Computer Vision and
  Pattern Recognition}. 49--59
\bibAnnoteFile{mueller2018ganerated}

\bibitem[{Newell et~al.(2016)Newell, Yang, and Deng}]{newell2016stacked}
Newell, A., Yang, K., and Deng, J. (2016).
\newblock Stacked hourglass networks for human pose estimation.
\newblock In \emph{European conference on computer vision} (Springer), 483--499
\bibAnnoteFile{newell2016stacked}

\bibitem[{Oberweger and Lepetit(2017)}]{oberweger2017deepprior++}
Oberweger, M. and Lepetit, V. (2017).
\newblock Deepprior++: Improving fast and accurate 3d hand pose estimation.
\newblock In \emph{Proceedings of the IEEE international conference on computer
  vision Workshops}. 585--594
\bibAnnoteFile{oberweger2017deepprior++}

\bibitem[{Oikonomidis et~al.(2011)Oikonomidis, Kyriazis, and
  Argyros}]{oikonomidis2011full}
Oikonomidis, I., Kyriazis, N., and Argyros, A.~A. (2011).
\newblock Full dof tracking of a hand interacting with an object by modeling
  occlusions and physical constraints.
\newblock In \emph{2011 International Conference on Computer Vision} (IEEE),
  2088--2095
\bibAnnoteFile{oikonomidis2011full}

\bibitem[{Qian et~al.(2020)Qian, Wang, Mueller, Bernard, Golyanik, and
  Theobalt}]{qian2020html}
Qian, N., Wang, J., Mueller, F., Bernard, F., Golyanik, V., and Theobalt, C.
  (2020).
\newblock Html: A parametric hand texture model for 3d hand reconstruction and
  personalization.
\newblock In \emph{European Conference on Computer Vision} (Springer), 54--71
\bibAnnoteFile{qian2020html}

\bibitem[{Rehg and Kanade(1994)}]{rehg1994digiteyes}
Rehg, J.~M. and Kanade, T. (1994).
\newblock Digiteyes: Vision-based hand tracking for human-computer interaction.
\newblock In \emph{Proceedings of 1994 IEEE Workshop on Motion of Non-rigid and
  Articulated Objects} (IEEE), 16--22
\bibAnnoteFile{rehg1994digiteyes}

\bibitem[{Romero et~al.(2017)Romero, Tzionas, and
  Black}]{MANO:SIGGRAPHASIA:2017}
Romero, J., Tzionas, D., and Black, M.~J. (2017).
\newblock Embodied hands: Modeling and capturing hands and bodies together.
\newblock \emph{ACM Transactions on Graphics, (Proc. SIGGRAPH Asia)} 36
\bibAnnoteFile{MANO:SIGGRAPHASIA:2017}

\bibitem[{Sandler et~al.(2018)Sandler, Howard, Zhu, Zhmoginov, and
  Chen}]{sandler2018mobilenetv2}
Sandler, M., Howard, A., Zhu, M., Zhmoginov, A., and Chen, L.-C. (2018).
\newblock Mobilenetv2: Inverted residuals and linear bottlenecks.
\newblock In \emph{Proceedings of the IEEE conference on computer vision and
  pattern recognition}. 4510--4520
\bibAnnoteFile{sandler2018mobilenetv2}

\bibitem[{Spurr et~al.(2018)Spurr, Song, Park, and Hilliges}]{spurr2018cross}
Spurr, A., Song, J., Park, S., and Hilliges, O. (2018).
\newblock Cross-modal deep variational hand pose estimation.
\newblock In \emph{Proceedings of the IEEE Conference on Computer Vision and
  Pattern Recognition}. 89--98
\bibAnnoteFile{spurr2018cross}

\bibitem[{Sridhar et~al.(2015)Sridhar, Feit, Theobalt, and
  Oulasvirta}]{sridhar2015investigating}
Sridhar, S., Feit, A.~M., Theobalt, C., and Oulasvirta, A. (2015).
\newblock Investigating the dexterity of multi-finger input for mid-air text
  entry.
\newblock In \emph{Proceedings of the 33rd Annual ACM Conference on Human
  Factors in Computing Systems}. 3643--3652
\bibAnnoteFile{sridhar2015investigating}

\bibitem[{Stenger et~al.(2006)Stenger, Thayananthan, Torr, and
  Cipolla}]{stenger2006model}
Stenger, B., Thayananthan, A., Torr, P.~H., and Cipolla, R. (2006).
\newblock Model-based hand tracking using a hierarchical bayesian filter.
\newblock \emph{IEEE transactions on pattern analysis and machine intelligence}
  28, 1372--1384
\bibAnnoteFile{stenger2006model}

\bibitem[{Sun et~al.(2019)Sun, Xiao, Liu, and Wang}]{sun2019deep}
Sun, K., Xiao, B., Liu, D., and Wang, J. (2019).
\newblock Deep high-resolution representation learning for human pose
  estimation.
\newblock In \emph{Proceedings of the IEEE/CVF Conference on Computer Vision
  and Pattern Recognition}. 5693--5703
\bibAnnoteFile{sun2019deep}

\bibitem[{Tan and Le(2019)}]{tan2019efficientnet}
Tan, M. and Le, Q. (2019).
\newblock Efficientnet: Rethinking model scaling for convolutional neural
  networks.
\newblock In \emph{International conference on machine learning} (PMLR),
  6105--6114
\bibAnnoteFile{tan2019efficientnet}

\bibitem[{Tang et~al.(2021)Tang, Wang, and Fu}]{tang2021towards}
Tang, X., Wang, T., and Fu, C.-W. (2021).
\newblock Towards accurate alignment in real-time 3d hand-mesh reconstruction.
\newblock In \emph{Proceedings of the IEEE/CVF International Conference on
  Computer Vision}. 11698--11707
\bibAnnoteFile{tang2021towards}

\bibitem[{Vaswani et~al.(2017)Vaswani, Shazeer, Parmar, Uszkoreit, Jones, Gomez
  et~al.}]{vaswani2017attention}
Vaswani, A., Shazeer, N., Parmar, N., Uszkoreit, J., Jones, L., Gomez, A.~N.,
  et~al. (2017).
\newblock Attention is all you need.
\newblock \emph{Advances in neural information processing systems} 30
\bibAnnoteFile{vaswani2017attention}

\bibitem[{Wan et~al.(2004)Wan, Luo, Gao, and Peng}]{wan2004realistic}
Wan, H., Luo, Y., Gao, S., and Peng, Q. (2004).
\newblock Realistic virtual hand modeling with applications for virtual
  grasping.
\newblock In \emph{Proceedings of the 2004 ACM SIGGRAPH international
  conference on Virtual Reality continuum and its applications in industry}.
  81--87
\bibAnnoteFile{wan2004realistic}

\bibitem[{Xiao et~al.(2018)Xiao, Wu, and Wei}]{xiao2018simple}
Xiao, B., Wu, H., and Wei, Y. (2018).
\newblock Simple baselines for human pose estimation and tracking.
\newblock In \emph{Proceedings of the European conference on computer vision
  (ECCV)}. 466--481
\bibAnnoteFile{xiao2018simple}

\bibitem[{Yang et~al.(2020)Yang, Li, Xu, Diao, and Lu}]{yang2020bihand}
Yang, L., Li, J., Xu, W., Diao, Y., and Lu, C. (2020).
\newblock Bihand: Recovering hand mesh with multi-stage bisected hourglass
  networks.
\newblock \emph{arXiv preprint arXiv:2008.05079}
\bibAnnoteFile{yang2020bihand}

\bibitem[{Zhang et~al.(2021)Zhang, Huang, Tan, Xu, Yang, Peng
  et~al.}]{zhang2021hand}
Zhang, X., Huang, H., Tan, J., Xu, H., Yang, C., Peng, G., et~al. (2021).
\newblock Hand image understanding via deep multi-task learning.
\newblock In \emph{Proceedings of the IEEE/CVF International Conference on
  Computer Vision}. 11281--11292
\bibAnnoteFile{zhang2021hand}

\bibitem[{Zhang et~al.(2019)Zhang, Li, Mo, Zhang, and Zheng}]{zhang2019end}
Zhang, X., Li, Q., Mo, H., Zhang, W., and Zheng, W. (2019).
\newblock End-to-end hand mesh recovery from a monocular rgb image.
\newblock In \emph{Proceedings of the IEEE/CVF International Conference on
  Computer Vision}. 2354--2364
\bibAnnoteFile{zhang2019end}

\bibitem[{Zhang et~al.(2018)Zhang, Zhou, Lin, and Sun}]{zhang2018shufflenet}
Zhang, X., Zhou, X., Lin, M., and Sun, J. (2018).
\newblock Shufflenet: An extremely efficient convolutional neural network for
  mobile devices.
\newblock In \emph{Proceedings of the IEEE conference on computer vision and
  pattern recognition}. 6848--6856
\bibAnnoteFile{zhang2018shufflenet}

\bibitem[{Zhou et~al.(2020)Zhou, Habermann, Xu, Habibie, Theobalt, and
  Xu}]{zhou2020monocular}
Zhou, Y., Habermann, M., Xu, W., Habibie, I., Theobalt, C., and Xu, F. (2020).
\newblock Monocular real-time hand shape and motion capture using multi-modal
  data.
\newblock In \emph{Proceedings of the IEEE/CVF Conference on Computer Vision
  and Pattern Recognition}. 5346--5355
\bibAnnoteFile{zhou2020monocular}

\bibitem[{Zimmermann and Brox(2017)}]{zimmermann2017learning}
Zimmermann, C. and Brox, T. (2017).
\newblock Learning to estimate 3d hand pose from single rgb images.
\newblock In \emph{Proceedings of the IEEE international conference on computer
  vision}. 4903--4911
\bibAnnoteFile{zimmermann2017learning}

\bibitem[{Zimmermann et~al.(2019)Zimmermann, Ceylan, Yang, Russell, Argus, and
  Brox}]{zimmermann2019freihand}
Zimmermann, C., Ceylan, D., Yang, J., Russell, B., Argus, M., and Brox, T.
  (2019).
\newblock Freihand: A dataset for markerless capture of hand pose and shape
  from single rgb images.
\newblock In \emph{Proceedings of the IEEE/CVF International Conference on
  Computer Vision}. 813--822
\bibAnnoteFile{zimmermann2019freihand}

\end{thebibliography}

\end{document}